\pdfoutput=1
\documentclass[11pt]{article}

\usepackage[]{acl}

\usepackage{times}
\usepackage{latexsym}
\usepackage{aligned-overset}
\usepackage{enumitem}

\usepackage{graphicx}
\graphicspath{{./img/}}
\usepackage{lipsum}

\usepackage{color}
\usepackage{xcolor}
\definecolor{newblue}{rgb}{0.0,0.0,0.7}

\usepackage[T1]{fontenc}

\usepackage[utf8]{inputenc}

\usepackage{microtype}
\usepackage{amssymb}
\usepackage{tabularx}
\usepackage{setspace}
\usepackage{amsmath}

\usepackage{xcolor}
\usepackage[linesnumbered,ruled,vlined]{algorithm2e}
\SetKwInput{KwInput}{Input}                
\SetKwInput{KwOutput}{Output}              

\newcommand{\ourtechnique}{PlanRAG}
\newcommand{\ourbenchmark}{DQA}

\title{\ourtechnique{}: A Plan-then-Retrieval Augmented Generation for Generative Large Language Models as Decision Makers}

\author{Myeonghwa Lee$^\ast$, Seonho An$^\ast$, Min-Soo Kim$^\dagger$ \\
        School of Computing, KAIST\\ \texttt{\{myeon9h, asho1, minsoo.k\}@kaist.ac.kr}}

\begin{document}
\maketitle
\begin{abstract}
In this paper, we conduct a study to utilize LLMs as a solution for decision making that requires complex data analysis. We define \textbf{Decision QA} as the task of answering the best decision, $d_{best}$, for a decision-making question $Q$, business rules $R$ and a database $D$. Since there is no benchmark that can examine Decision QA, we propose Decision QA benchmark, \textbf{DQA}. It has two scenarios, Locating and Building, constructed from two video games (Europa Universalis IV and Victoria 3) that have almost the same goal as Decision QA. To address Decision QA effectively, we also propose a new RAG technique called the \textit{iterative plan-then-retrieval augmented generation} (\textbf{PlanRAG}). Our PlanRAG-based LM generates the plan for decision making as the first step, and the retriever generates the queries for data analysis as the second step. The proposed method outperforms the state-of-the-art iterative RAG method by 15.8\% in the Locating scenario and by 7.4\% in the Building scenario, respectively. We release our code and benchmark at \url{https://github.com/myeon9h/PlanRAG}.
\end{abstract}

\def\thefootnote{$\ast$}\footnotetext{Equal contribution.}
\def\thefootnote{$\dagger$}\footnotetext{Corresponding author.}
\renewcommand{\thefootnote}{\arabic{footnote}}

\section{Introduction}
\label{sec:introduction}

In many business situations, decision making plays a crucial role for the success of organizations \cite{kasie2017decision, gupta2002implementing}.
Here, decision making involves analyzing data, ultimately leading to the selection of the most suitable alternative to achieve a specific goal \cite{provost2013data, divan2017data}.
For example, we assume that one of the goals of the pharmacy company “Pfizer” is to minimize the production cost while maintaining on-time delivery from plants to customers in the pharmaceutical distribution network \cite{gupta2002implementing}, and the production cost is proportional to the amount of operation time and number of employees of a plant. Then, Pfizer may face the following decision-making problems: (P1) which plant it should operate or stop, and (P2) how many employees it should hire for each plant.

In general, the decision-making task requires performing the following three steps: (1) making a plan for which kind of analysis is needed for decision; (2) retrieving necessary data using queries; (3) making a decision (i.e., answering) based on the data \cite{troisi2020growth, sala2022data}.
To make the Steps (2) and (3) easier, a lot of decision support systems have been developed and utilized during the past few decades \cite{gupta2002implementing, eom2006survey, power2007brief, hedgebeth2007data, power2008understanding, kasie2017decision}.
However, humans still have been in charge of the most hard part, Step (1).
The goal of this study is to investigate the possibility of replacing the human role with a Large Language Model(LLM) such that it performs not only Steps (2) and (3) but also Step (1), that is, all the Steps end-to-end.

\begin{figure*}[t!]
    \centering
    \includegraphics[width=\textwidth]{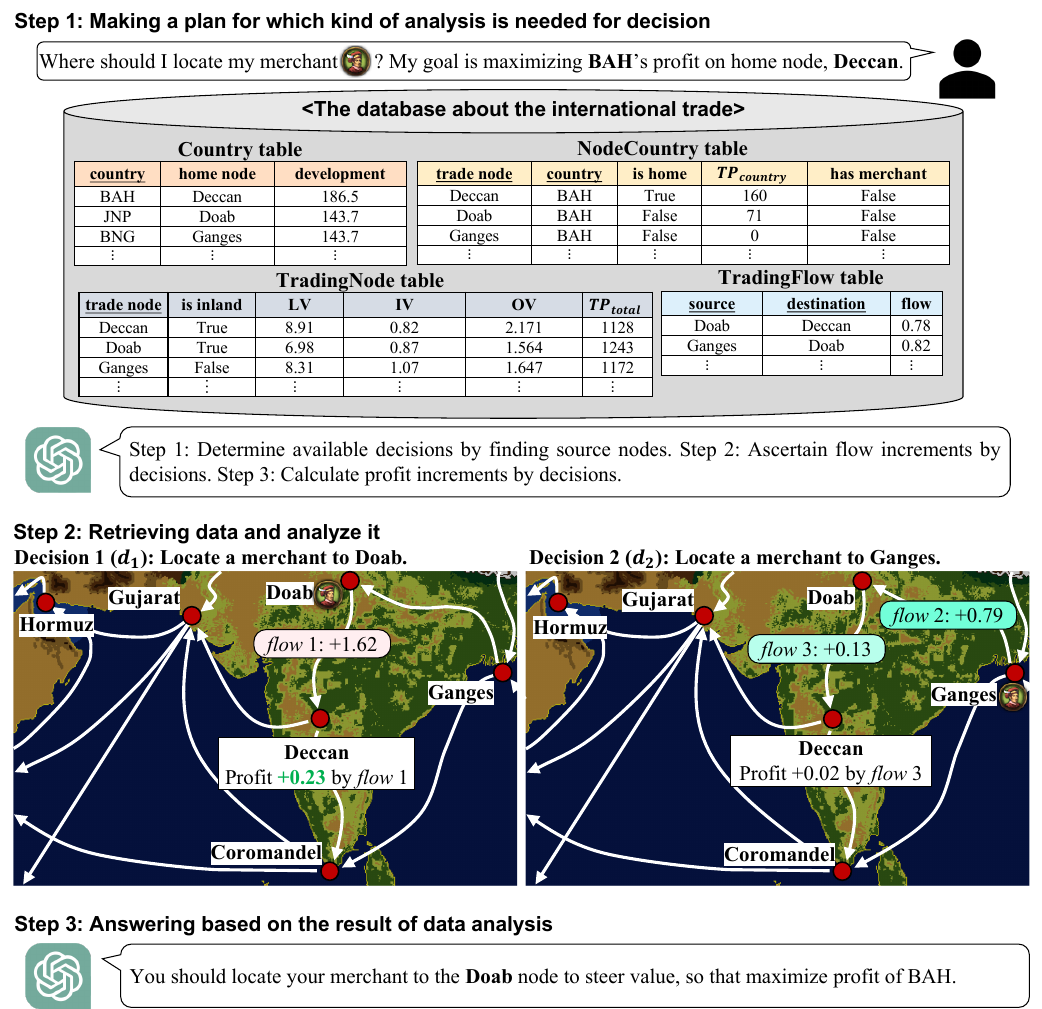}
    \caption{Example of Decision QA. 
    A red dot represents a trading node. A \textit{Profit} in the Deccan box indicates a potential profit change by each decision.
    Note that the potential profit changes are not in the database, which should be calculated from the database.
    Each country has only a single main(home) trading node. The \underline{underlined} column names in a table indicate the key of the table.
    }
    \label{fig:example}
\end{figure*}

To achieve the goal, we propose, \textbf{Decision QA}, a new decision-making task for language models.
Decision QA is defined as a QA-style task that takes a pair of database $D$, business rules $R$ and a decision-making question $Q$ as input and generates the best decision as output.
Figure \ref{fig:example} shows a situation in Europa Universalis IV game where countries compete in trade at the Age of Discovery, as an example of Decision QA. 
Each country decides which trading city (i.e., node) it should locate a merchant on to maximize its profit on its main (i.e., home) trading node.
The example shows that a decision-making LLM decides to locate a merchant in Doab to maximize the profit of Deccan, the home trading node of the country BAH, after analyzing the database about international trade.

Next, we propose a benchmark for Decision QA called \textbf{\ourbenchmark{}}, which consists of the following two scenarios: Locating and Building. 
The former scenario consists of decision questions like “Which trade node should I locate a merchant on" (similar to P1 of Pfizer). The latter consists of questions like “How many woods should I supply to a factory” (similar to P2 of Pfizer).
Due to the difficulty of building \ourbenchmark{} using real-world business data, we built the benchmark by extracting the game data involving 301 specific situations from the two video games Europa Universalis IV and Victoria 3\footnote{Grand strategy games published by \href{https://www.paradoxinteractive.com/}{Paradox Interactive}}, which well imitate real business situations.
To eliminate the randomness of the game and publish our benchmark, we develop the game simulators that record the decision results for 301 situations.
We utilize the results as annotations for the questions of \ourbenchmark{}.

The recent breakthrough in LLMs is making it possible to replace Steps (2) and (3) of the decision making task with LLMs, in particular, based on \textit{Retrieval-Augmented Generation} (RAG) technique. So far, a lot of RAG-based methods have been proposed for various tasks\cite{lewis2020retrieval, DBLP:conf/iclr/KhandelwalLJZL20, izacard-grave-2021-leveraging, DBLP:conf/icml/BorgeaudMHCRM0L22, JMLR:v24:23-0037, yasunaga2022retrieval, jiang-etal-2022-retrieval, shi2023replug}.
In these methods, a retriever finds external data highly relevant to a question and conveys it to LMs, so that LMs can generate an answer based on it \cite{lewis2020retrieval}.
Recently, the \textit{iterative RAG} technique has also been proposed to address more complex problems that should utilize retrieved results to perform further retrievals \cite{trivedi-etal-2023-interleaving, jiang2023active}.

However, the existing RAG-based methods mainly focus on the knowledge-based QA tasks \cite{karpukhin-etal-2020-dense, trivedi-etal-2023-interleaving}, but do not focus on the decision making QA task. 
As a result, they are not very good at handling Step (1), i.e., making a plan for decision, in our observation.
For instance, in Figure \ref{fig:example}, an LM for decision making should reason which analysis is needed to perform for maximize the profit of Deccan. However, the existing methods just try to identify, for example, what Deccan is.

To address this limitation, we propose the iterative plan-then-retrieval augmented generation technique, \textbf{\ourtechnique{}}, which extends the iterative RAG technique for Decision QA.
A PlanRAG-based LM first makes a plan for which kind of analysis it needs by examining data schema and questions (the \textit{planning} step).
Next, it retrieves the scattered pieces of data for the analysis by generating and posing queries (the \textit{retrieving} step).
Finally, it assesses whether it needs to make a new plan for further analysis, and then repeats both the planning and retrieval steps iteratively (the \textit{re-planning} step), or makes a decision based on the data (the \textit{answering} step).

To validate the effectiveness of our \ourtechnique{} on Decision QA, we compare both the state-of-the-art iterative RAG-based LM and our \ourtechnique{}-based LM for the \ourbenchmark{} benchmark.
Our contributions are summarized as follows:

\begin{itemize}
    \item We define a new challenging task, \textbf{Decision QA}, which requires both planning and data analysis for decision making.
    \item We propose the benchmark for Decision QA called \textbf{\ourbenchmark{}} having two scenarios of Locating and Building.
    \item We propose a new retrieval-augmented generation technique, \textbf{\ourtechnique{}}, that can significantly enhance the capability of decision making of LLMs. 
    \item We demonstrated that PlanRAG is far more effective than the iterative RAG technique for Decision QA.
\end{itemize}

\section{Decision QA task}
\label{sec:decisionQAtask}

We define \textbf{Decision QA} as the task of answering the best decision $d_{best}$ given a decision-making question $Q$, business rules $R$, and structured database $D$ that follows a schema $S$.
Here, $Q$ contains a textual goal that users want to be achieved through the decision $d_{best}$, and $R$ contains textual descriptions of formulas that are referenced to reason $d_{best}$.

Without loss of generality, the database $D$ is too large to fit in the input of an LM all at once. 
Therefore, we assume that an LM retrieves data from $D$ by posing a query for data analysis, which we call as a \textit{data analysis query} hereafter. 

We assume that $D$ is either a \textit{labeled property graph (LPG) database} or a \textit{relational database (RDB)}. For example, the database in Figure \ref{fig:example} is a relational database. Here, an LPG refers to a graph with properties on edges and nodes, which is widely used in the industry \cite{akoglu2015graph, guo2020survey}. We call a labeled property graph database as simply a \textit{graph database}, or \textit{GDB} hereafter.

\section{\ourbenchmark{}: Decision QA benchmark}
\label{sec:ourdataset}

\subsection{Locating scenario}
\label{subsec:locating}

The database in this scenario is composed of the following four tables when $D$ is RDB. 
It also can be easily represented as GDB by regarding the tuples in TradingFlow and NodeCountry as edges and the tuples in TradingNode and Country as vertices.

\begin{table}[htb!]
\centering
\includegraphics[width=\columnwidth]{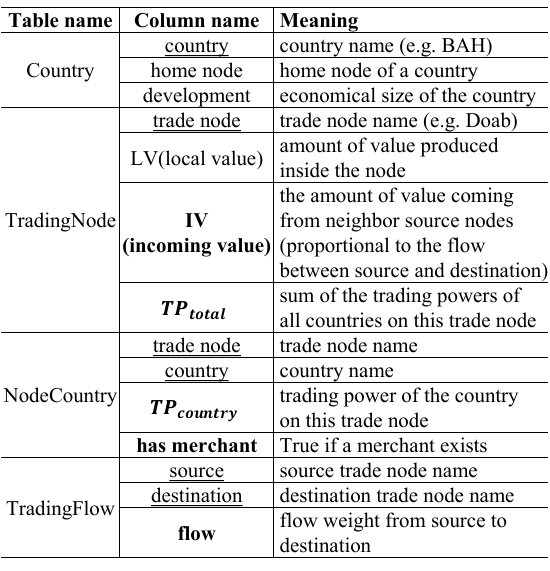}
\caption{RDB schema for Locating. The \textbf{bold} indicate the columns which values depend on other values and a user decision. Some columns are omitted.
}
\label{tab:locating}
\end{table}

There are some business rules for decision making. The column values in bold are calculated from some other values in the database and a user decision (e.g., location of a merchant) according to the rules. If a user locates a merchant on a trading node for a country, the flow from the trading node to the home node of the country increases.
In the below rules, $c$ is a country, $n$ a trading node, \textit{src} a source node, \textit{dest} a destination node, and $h$ the home node of $c$.
We also denote the set of countries as $C$, and the set of TradingFlow tuples as $F$. TPR means Trading Power Ratio.

\begin{align}
    & TPR(n,c) = TP_{\textit{country}}(n,c)/TP_{\textit{total}}(n) \label{eq:TPR} \\
    & IV(\textit{dest})=1.05 \cdot \Sigma_{(\textit{src},\textit{dest}) \in \text{F}} \textit{flow}(\textit{src},\textit{dest})\label{eq:IV} \\
    & \textit{profit}(c) = (LV(h)+IV(h)) \cdot TPR(h,c)\label{eq:profit}
\end{align}

In this scenario, the goal of decision making is to choose trading node that can maximize $\Delta \textit{profit}(c)$ of the given target country $c$.

\subsection{Building scenario}
\label{subsec:building}
The database $D$ in this scenario is composed of the following four tables when $D$ is RDB. 
It also can be easily represented as GDB by regarding the tuples in Demand and Supply as edges and the tuples in Goods and Buildings as vertices.

\begin{table}[htb!]
\centering
 \includegraphics[width=\columnwidth]{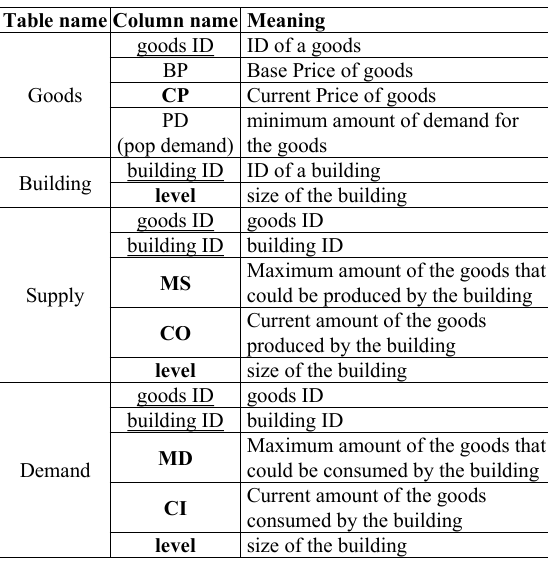}
\caption{
RDB schema for Building. Some columns are omitted.
}
\label{tab:building}
\end{table}

\begin{figure}[htb!]
    \centering
    \includegraphics[width=\columnwidth]{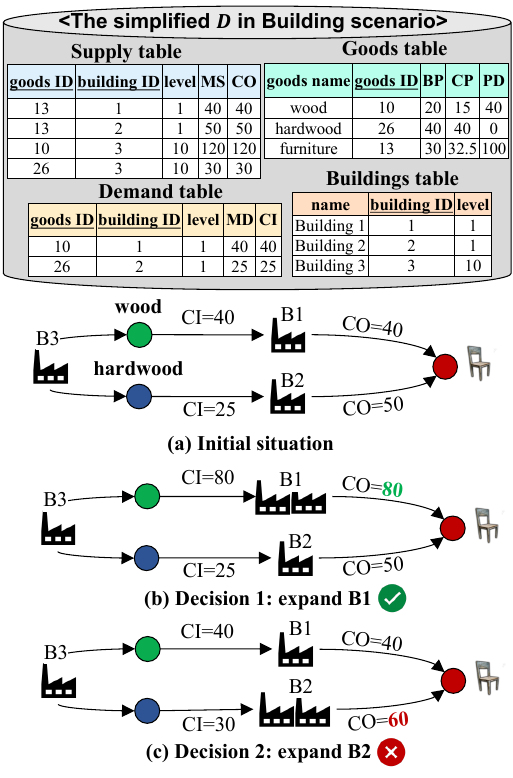}
    \caption{
    Example of the Building scenario. The red, green, and blue circles indicate furniture, wood, and hardwood, respectively. We assume that the goal is reducing the price of furniture by deciding the factory between B1 and B2 to expand and so increase its production.
    }
    \label{fig:building-scenario}
\end{figure}
The most basic business rule is that $CO(g,b)$ increases for any goods $g$, if a decision maker expands the factory building $b$.
In the below rules, $g$ is goods, and $b$ is a building. We also denote the set of Supply as \textit{Sup}, and the set of Demand as \textit{Dem}. $TD$ means Total Demand, and $TS$ Total Supply.
Other major business rules are as follow:

\begin{align}
    {TD}(g) &= PD(g) + \Sigma_{(g,b) \in \textit{Dem}} MD(g,b)\label{eq:TD}\\
    {TS}(g) &= \Sigma_{(g,b) \in \textit{Sup}} CO(g,b) \label{eq:TS}\\
    CP(g) &=  BP(g) \cdot \notag \\
    & (1+0.75\cdot\frac{{TD}(g)-{TS}(g)}{\textit{max}({TD}(g), {TS}(g))}) \label{eq:CP}
\end{align}

In this scenario, the goal is to minimize $CP(g)$ for a given goods $g$.

\subsection{Statistics of \ourbenchmark{}}
\label{subsec:dataset-statistics}

\ourbenchmark{} consists of a total of 301 pairs of $D$ and $Q$: (1) 200 pairs for the Locating scenario and (2) 101 pairs for the Building scenario. 
Each database again has two versions, RDB and GDB, for the same corresponding question, and so, a total of 602 databases are provided in the \ourbenchmark{} benchmark. 
We assume that SQL is used for RDB, while Cypher Query Language (CQL) \cite{francis2018cypher} used for GDB.
Table \ref{tab:stats} shows some statistics of the databases in \ourbenchmark{}.
Details for data collection are in Appendix \ref{appendix:data-coll}.

\begin{table}[htb!]
\renewcommand{\arraystretch}{1.1}
\footnotesize
\centering
\begin{tabular}{lrr}
\hline
\textbf{Statistics} & \textbf{Locating} & \textbf{Building}\\
\hline
\# of $\langle$$Q$,$D$$\rangle$ pairs & 200 & 101 \\
\hline
\textbf{Relational DB (RDB)} &  & \\
Avg. rows in tables of $D$ & 2038.8 & 579.0 \\ 
Avg. cols in tables of $D$ & 4.5 & 4.5\\
\hline
\textbf{Graph DB (GDB)} &  & \\
Avg. \# of edges of $D$ & 1432.3 &  374.7 \\ 
Avg. \# of vertices of $D$ & 606.5 & 204.3 \\
\hline
\end{tabular}
\caption{Statistics about the databases in \ourbenchmark{}.
}

\label{tab:stats}
\end{table}

\section{Methodology: PlanRAG}
\label{sec:methodology}

For Decision QA, the existing RAG technique \cite{jiang2023active, trivedi-etal-2023-interleaving} tries to answer the best decision $d_{best}$ for given $\langle$$Q$,$S$,$R$$\rangle$ through a single type of reasoning that utilizes on the results retrieved from $D$ by data analysis queries. Figure 3 (a) shows its reasoning process. 
If the retrieval from $D$ is performed only once, the process is called as \textit{single-turn RAG}.
Otherwise, if the retrieval is performed multiple times, the process is called as \textit{iterative RAG}.

\begin{figure*}[htb!]
    \centering
    \includegraphics[width=\textwidth]{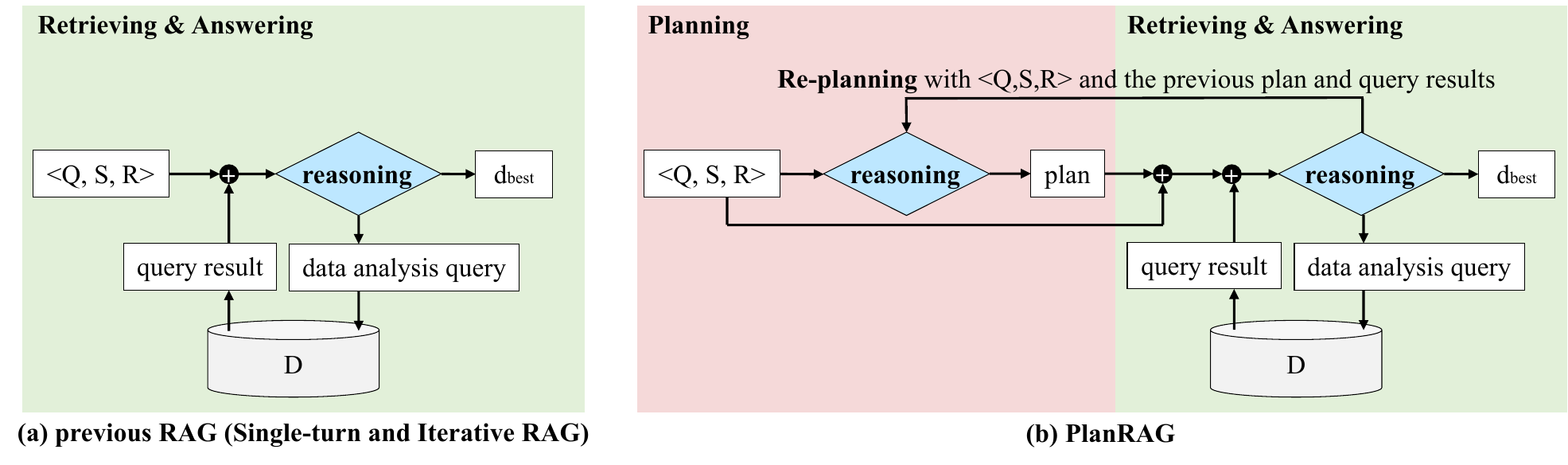}
    \caption{Comparison of the reasoning processes of between previous RAG and our PlanRAG techniques.
    }
    \label{fig:reasoning-process}
\end{figure*}

In contrast, our \textit{iterative plan-then-retrieval augmented generation} (PlanRAG) technique tries to answer $d_{\textit{best}}$ through two types of reasoning. 
The first type of reasoning is making a plan, and the second type of reasoning is similar to the reasoning of the existing RAG, i.e., answering based on the results retrieved from $D$ by of data analysis queries.
In particular, we built a single LM that can perform both types of reasoning because to reduce the side effects of using separate LMs.
We prompted the LM by adding the `Plan' and `Re-plan' instructions to ReAct \cite{yao2022react}, about which details are in Appendix. \ref{appendix:prompt-planrag}.
Figure \ref{fig:reasoning-process} (b) shows the reasoning process of PlanRAG, where we explain the steps of (1) planning, (2) retrieving \& answering, and (3) re-planning, in detail as follows.

\begin{figure*}[ht!]
    \centering
    \includegraphics[width=\textwidth]{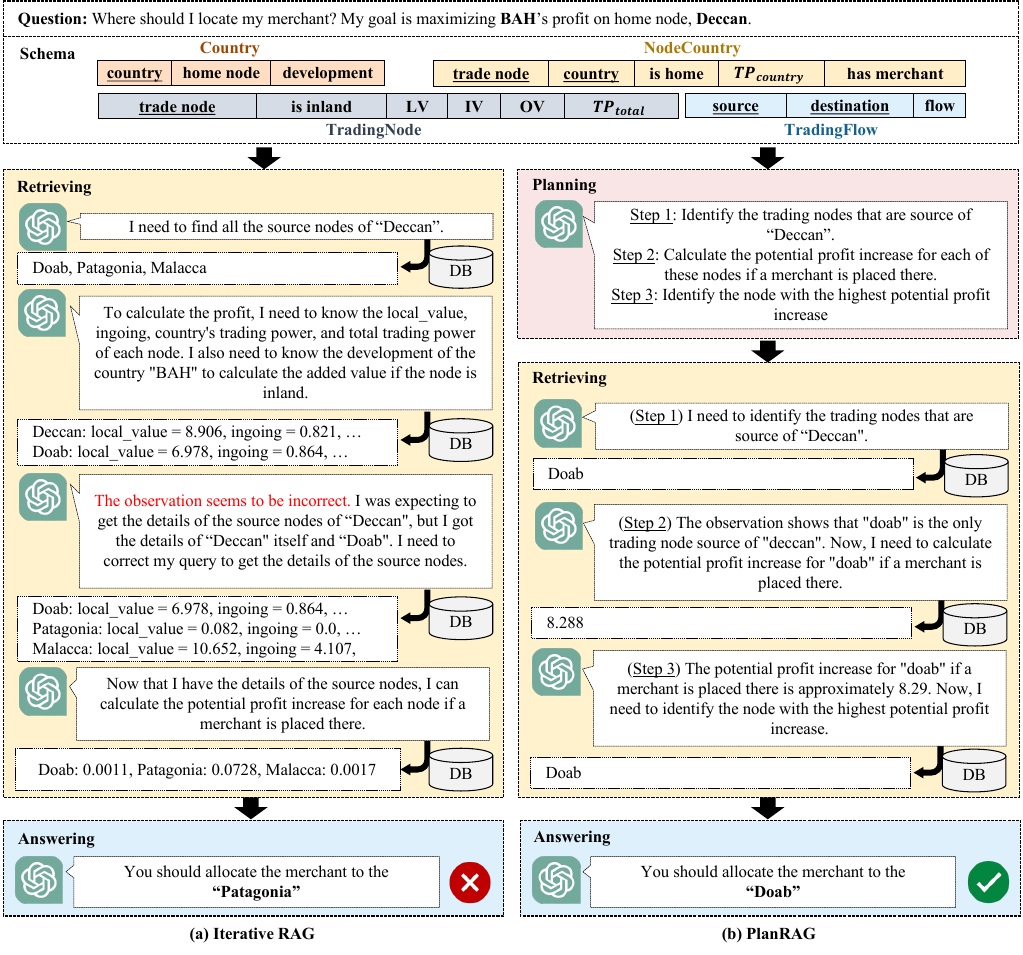}
    \caption{Example of the reasoning processes of previous Iterative RAG and our PlanRAG in the Locating scenario.}
    \label{fig:dataset-example}
\end{figure*}

\textbf{Planning:} In this step, an LM takes $\langle$$Q$, $S$, $R$$\rangle$ as input, and then generates an initial plan for data analysis.
The initial plan describes a series of data analyses that needs for decision making and so need to be performed in the retrieving step.
The red box in Figure \ref{fig:dataset-example}(b) shows its example.

\textbf{Retrieving \& Answering:} Unlike the previous RAG technique, an LM takes not only $\langle$$Q$, $S$, $R$$\rangle$ but also the initial plan as input.
Then, it can generate data analysis queries for decision making much more effectively than the previous RAG. 
Figure \ref{fig:dataset-example} shows how PlanRAG-based LM generates the queries differently from the previous RAG. 
The queries are actually executed by SQL or Cypher to the database through RAG interfaces such as LangChain\footnote{\href{https://langchain.readthedocs.io/en/latest}{https://langchain.readthedocs.io/en/latest}} 
and LlamaIndex\footnote{\href{https://docs.llamaindex.ai/en/stable/}{https://docs.llamaindex.ai/en/stable/}}.
The query results are used iteratively for reasoning about whether it needs re-planning or just a further retrieval for better decision making.
Through the backward link to the planning process, the planning and retrieving processes are iteratively performed until an LM determines that there is no longer a need for further analysis to make a decision.

\textbf{Re-planning:} Re-planning is done when the initial plan is not good enough to solve the decision question.
In order to make the LM possible to decide whether re-planning or not, we prompted the LM with some instructions to assess the current plan by referring to the result of each retrieval step (see Appendix. \ref{appendix:prompt-planrag} for details).
As a result, the LM takes not only $\langle$$Q$, $S$, $R$$\rangle$ but also a current plan and query results as input and generates a new plan to do further analysis, or correct the direction of previous analysis.

\section{Experiments}
\label{sec:experiments}

\subsection{Experimental Setup}
\label{subsec:setup}

To validate the effectiveness of the proposed PlanRAG for the Decision QA task,
we implement and compare four different decision-making LMs: (1) SingleRAG-LM based on Single-turn RAG, (2) IterRAG-LM based on Iterative RAG, (3) PlanRAG-LM based on PlanRAG, (4) PlanRAG-LM w/o RP, which means PlanRAG without re-planning.
Prompts based on ReAct \cite{yao2022react} are used to make these LMs as decision makers (see details in Appendix \ref{appendix:prompt}).
All these decision makers are implemented by GPT-4 \cite{openai2023gpt} with a zero temperature and the LangChain library.
In terms of database, we use MySQL\footnote{\href{https://www.mysql.com}{https://www.mysql.com}} for an RDBMS and Neo4j\footnote{\href{https://neo4j.com}{https://neo4j.com}} for a GDBMS.

All experiments are conducted in a zero-shot and single run setting.
This zero-shot setting is predicated on the following two reasons. 
First, in most of real-world business situations, it is very hard to know the strategies for making the best decisions in advance. 
Second, in few-shot settings, we have observed that an LM is overfitted to not only the problem-solving strategy but also the content of a database in the given shots.

An answer from a decision maker is considered correct only if it is semantically identical to a ground-truth best decision on \ourbenchmark{}. 
For example, in Figure \ref{fig:dataset-example} (b), we consider the answer is correct since the ground-truth best decision is \textit{Doab}.

\subsection{Results and Analysis}
\label{subsec:results}
\noindent \textbf{Main results:}
Table \ref{tab:results} presents the experimental results, where 
significantly improves the performance of decision making for both scenarios, by 15.8\% for Locating and 7.4\% for Building, compared to the existing SOTA technique, Iterative RAG \cite{yao2022react}.
It shows well the effectiveness of PlanRAG for the decision making task. 
The reason why PlanRAG is relatively more effective in Locating than in Building is that 
Building requires a longer traversal than Locating scenario and it makes planning harder than Locating.
The accuracy of SingleRAG-LM in Building is very low, which is because the Building scenario requires generating a very complex query that is hard to be reasoned at once.
SingleRAG-LM failed to retrieve any results from the database in over 60\% of Locating and 95\% of Building questions.
Table \ref{tab:results} also present that no re-planning in PlanRAG leads to a decrease in accuracy, in particular, by 10.8\% in Locating and 0.9\% in Building.
This result shows that the re-planning process is helpful and important to the decision maker LM of the PlanRAG technique for the decision making task.

\begin{table}[htb!]
\renewcommand{\arraystretch}{1.25}
\footnotesize
\centering
\begin{tabular}{lcc}
\hline
\textbf{Decision makers}&\textbf{Locating} & \textbf{Building}\\
\hline
\textbf{Single-turn RAG} & & \\
SingleRAG-LM & 30.5 & 2.5 \\
\hline
\textbf{Iterative RAG} & & \\
IterRAG-LM & 48.5 & 37.6 \\
\hline
\textbf{\ourtechnique{} (ours)} & & \\
PlanRAG-LM & \textbf{64.3} & \textbf{45.0}\\
PlanRAG-LM w/o RP & 53.5 & 44.1 \\ 
\hline
\end{tabular}
\caption{Accuracy(\%) of the techniques for DQA (Each accuracy is an average of the accuracies in RDB and GDB).}
\label{tab:results}
\end{table}

\begin{figure}[htb!]
\centering
    \includegraphics[width=\columnwidth]{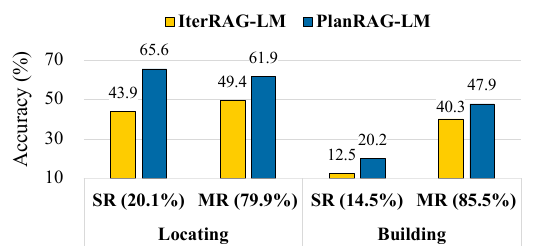}
    \caption{
    Accuracy(\%) of IterRAG-LM and PlanRAG-LM for SR and MR questions.
    }
    \label{fig:srmr}
\end{figure}

\noindent \textbf{Analysis for SR and MR:}
There are relatively simple questions and relatively hard questions in DQA. To check the effectiveness of PlanRAG according to the degree of difficulty of questions, we divide \ourbenchmark{} questions into two different types: (1) Single retrieval (\textit{SR}) questions, and (2) Multiple Retrieval (\textit{MR}) questions.
Here, SR refers to the case where IterRAG-LM performs a single retrieval from $D$ for solving the question, while MR refers to the questions that it performs multiple retrievals for solving the question. 
There are a total of 84 $\langle$$Q$, $D$$\rangle$ pairs of SR and 518 $\langle$$Q$, $D$$\rangle$ pairs of MR. We compare the accuracy of IterRAG-LM and PlanRAG-LM for SR and MR, which results are presented in Figure \ref{fig:srmr}.

In the result, PlanRAG-LM outperforms much more IterRAG-LM for the SR questions than for the MR questions. 
It is because the SR questions actually are not so easy in many cases.  
They are the ones that IterRAG-LM tried to solve only by using a single retrieval. 
That is, some of them are the questions that IterRAG-LM underestimated its degree of difficulty, but actually are relatively hard ones that require multiple retrievals.
In contrast, PlanRAG-LM reduces the likelihood that it understands the degree of difficulty of given questions through the planning step and performs multiple retrievals according to the plan. As a result, it could significantly improve the accuracy.
For the MR cases, PlanRAG-LM is still more effective than IterRAG-LM because the former performs data retrievals relatively systematically according to the plan, whereas the latter performs retrievals relatively in a disorganized manner like in Figure \ref{fig:dataset-example} (a).

\noindent \textbf{Analysis for RDB and GDB:}
Table \ref{tab:results-df} presents the accuracy of LMs for two different databases, RDB and GDB, in DQA.
In the results, PlanRAG-LM is more effective than other LMs in both scenarios regardless of the database types, i.e., RDB and GDB.
We note that PlanRAG-LM is more effective for GDB than for RDB in the Building scenario.
It is because Building is a harder scenario than Locating that requires a longer traversal in GDB (or a more number of joins in RDB) for answering a question. 
For example, the questions in Locating need just a single-hop traversal from source nodes to the home node, but the ones in Building need a multi-hop traversal to find high supply goods in Figure \ref{fig:building-scenario}.

\begin{table}[h]
\footnotesize
\centering
\renewcommand{\arraystretch}{1.25}
\begin{tabular}{lcccc}
\hline
& \multicolumn{2}{c}{\textbf{Locating}} & \multicolumn{2}{c}{\textbf{Building}} \\ \cline{2-5} 
\textbf{Decision makers} & RDB & GDB & RDB & GDB \\ 
\hline
SingleRAG-LM & 25.5 & 35.5 & 2.0 & 3.0 \\
IterRAG-LM & 37.5 & 59.5 & 34.7 & 40.6 \\
PlanRAG-LM & \textbf{64.5} & \textbf{64.0} &\textbf{40.6} & \textbf{49.5} \\ 
\hline
\end{tabular}
\caption{Accuracy(\%) of LMs for RDB and GDB.}
\label{tab:results-df}
\end{table}

\noindent \textbf{Rate of missed data analysis:}
In each scenario, there are some critical values that need to be queries or calculated in order to answer the questions. 
For example, Locating has such values including $IV$ and $TP_{\textit{total}}$, and Building has such values including $CO$ and $PD$. In order to analyze why PlanRAG is more effective than IterRAG-LM, we measure the rate of missed data analysis for querying or calculating such values. We use $IV$ and $TP_{\textit{total}}$ for Locating, and $CO$ and $PD$ for Building, as a criteria. 
Table \ref{tab:results-da} shows that PlanRAG-LM has low rates, 1.3\% and 21.8\%, while IterRAG-LM has higher rates, 3.3\% and 33.2\%. It means that IterRAG-LM would achieve lower accuracy than PlanRAG-LM even though it could do reasoning perfectly. According to Table \ref{tab:results-da}, PlanRAG-LM might be able to achieve the accuracy of 98.7\% in Locating, but the actual accuracy in Table \ref{tab:results} is lower than that. It is because reasoning (including planning) itself is very challenging besides missed data analysis. 

\begin{table}[h]
\centering
\begin{tabular}{lcc}
\hline
    \textbf{Decision makers} & \textbf{Locating} & \textbf{Building} \\ \hline
    IterRAG-LM               & 3.3\% & 33.2\%      \\
    PlanRAG-LM               & \textbf{1.3\%}   & \textbf{21.8\%}     \\ \hline
\end{tabular}
\caption{Rate of missed data analysis. }
\label{tab:results-da}
\end{table}

\begin{figure}[htb!]
\centering
    \includegraphics[width=\columnwidth]{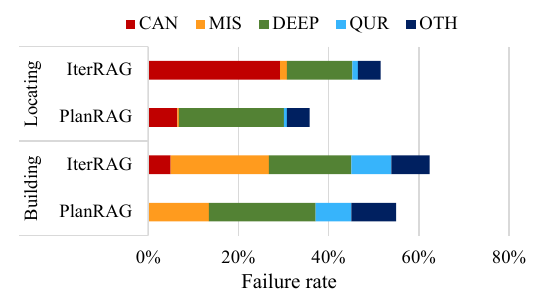}
    \caption{A result of failure case analysis. Failure rate means the number of failure questions divided by the number of all questions.
    }
    \label{fig:failure-case}
\end{figure}

\noindent \textbf{Analysis for failure cases:}
We classify each failure case into five error categories as follows: (1) \textit{CAN}, which means an LM fails to solve a question by considering improper candidates (e.g. \textit{dest} of Deccan in Figure \ref{fig:example}) and answer them; (2) \textit{MIS}, missed data analysis; (3) \textit{DEEP}, using retrieved data or equations unproperly; (4) \textit{QUR}, query generation error; and (5) \textit{OTH}, other errors (e.g. exceeding token length limits). For example, we classify \ref{fig:dataset-example}(a) as DEEP, because an LM misused some equations and so underestimated the profit of Doab. 
We compare failure cases done by IterRAG-LM and PlanRAG-LM based on these categories, in Figure \ref{fig:failure-case}. 

In the result, PlanRAG-LM significantly reduces CAN and MIS errors for both scenarios.
It means that PlanRAG-LM can understand a question of Decision QA better and query critical data for the question better than IterRAG-LM.
We also note that PlanRAG-LM has slightly more DEEP cases than IterRAG-LM for both scenarios.
In our observation, DEEP error appears only if there are no CAN or MIS errors.
For example, in Figure \ref{fig:dataset-example}(a), there is no chance for an LM to underestimate the profit of Doab (i.e. making DEEP error) when it fails to take Doab as one of the candidates (i.e. making CAN error).
Thus, we can say that the increase of DEEP cases in PlanRAG-LM comes from side effects of reducing both CAN and MIS cases.

\noindent \textbf{Analysis for re-planning:}
PlanRAG-LM performs re-planning for some DQA questions.
Table \ref{tab:replan-analysis} presents the distribution of questions based on the number of re-plannings conducted by PlanRAG-LM and the accuracy improved by re-planning.
Detailed cases and statistics for re-planning are in Appendix \ref{appendix:replan-case}.

\begin{table}[htb!]
\renewcommand{\arraystretch}{1.2}
\footnotesize
\centering
    \begin{tabular}{lcc}
    \hline
    \textbf{No. of re-plannings} & \textbf{Locating} & \textbf{Building} \\ \hline
    0 & 376 (66.8\%) & 125 (57.6\%) \\
    1 & 0 (-) & 0 (-) \\
    2 & 12 (41.7\%) & 24 (41.7\%) \\
    3 & 5 (20.0\%) & 23 (21.7\%) \\
    more than 4 & 7 (0.0\%) & 30 (13.3\%) \\ \hline
    Total & 400 (64.3\%) & 202 (45.0\%) \\ \hline
    \end{tabular}
\caption{Statistics of DQA questions according to the number of re-plannings conducted by PlanRAG-LM. The percentage in parentheses indicates the accuracy(\%) on a set of corresponding questions.}
\label{tab:replan-analysis}
\end{table}

In the result, PlanRAG-LM re-plans more frequently in the Building scenario than the Locating scenario.
PlanRAG-LM re-plans 24 out of 400 questions (6\% of questions) in Locating, but re-plan 77 out of 202 questions (38\% of questions) in Building.
In addition, the rate of questions that are re-planned more than four times is much larger in Building (30 questions, 14.85\%) than Locating (7 ones, 1.75\%). 
We note that an LM re-plans if its original plan is insufficient, as we explained in Section \ref{sec:methodology}.
Thus, the result indicates that an LM struggles to make a plan (for both initial planning and re-planning) in Building and explains relatively small gap between PlanRAG-LM and other techniques in Building, in Table \ref{tab:results}. 
It is also consistent with the result in Table 7 where the accuracy improved by re-planning decreases as the number of re-plannings increases.

\color{black}

\section{Related Work}
\label{sec:related-work}
\noindent \textbf{NLP tasks using structured data}
A number of benchmarks for reasoning over structured data have been proposed, such as Table NLI benchmarks \cite{DBLP:conf/iclr/ChenWCZWLZW20, gupta-etal-2020-infotabs, jena-etal-2022-leveraging} and Tabular QA \cite{iyyer-etal-2017-search, chen-etal-2020-hybridqa, zhu-etal-2021-tat, chen-etal-2021-finqa, chen2021ottqa, li-etal-2022-learning, nan-etal-2022-fetaqa}. 
Tabular QA is a task answering questions based on given tabular data, and Tabular NLI is a task determining hypotheses are entailment, contradiction, or neutral based on given tabular data.
However, such benchmarks did not consider the business rules, and also, not consider LMs' querying over a large structured database.
Table \ref{tab:num-of-rows} shows that the number of rows per table in the above benchmarks is much smaller than that in our DQA benchmark.

\begin{table}[h]
\renewcommand{\arraystretch}{1.2}
\footnotesize
\centering
\begin{tabular}{lr}
\hline
\textbf{Benchmarks} & \textbf{Avg. rows}\\
\hline
\textbf{Tabular QA} &\\
HybridQA \cite{chen-etal-2020-hybridqa} & 15.7\\
TAT-QA \cite{zhu-etal-2021-tat} & 9.5\\
FinQA \cite{chen-etal-2021-finqa} & 6.26\\
WikiTableQA \cite{zhu-etal-2021-tat} & 30\\
\hline
\textbf{Table NLI} & \\
TabFact \cite{DBLP:conf/iclr/ChenWCZWLZW20} & 14.5\\ 
ToTTo-TNLI \cite{jena-etal-2022-leveraging} & 35.8\\
\hline
\textbf{Decision QA (ours)} & \\
Building with RDB & 579.0 \\
Locating with RDB & \textbf{2038.8}\\
\hline
\end{tabular}
\caption{Average number of rows per table in Tabular QA, Table NLI and Decision QA benchmarks. For Tabular QA and Table NLI, and DQA benchmarks.
For DQA, the number of rows of $D$ are presented since every question needs to access to all tables.}
\label{tab:num-of-rows}
\end{table}

\noindent \textbf{RAG technique}
RAG is the most common approach to augment the generation of LMs with external data.
LMs retrieve the data related to an input (e.g., question) and then, generate a response (e.g., answer) based on the retrieved observations. Most of them operate in a single-turn (i.e., non-iterative) manner \cite{guu2020retrieval, JMLR:v24:23-0037, izacard-grave-2021-leveraging, jiang2022retrieval, shi2023replug, DBLP:conf/icml/BorgeaudMHCRM0L22, lewis2020retrieval} and so have clear limitations in complex tasks that require multi-hop reasoning. 
To address this, several methods have been proposed to augment generation by performing a process of retrieval-then-generation iteratively \cite{jiang2023active, shao2023enhancing, trivedi-etal-2023-interleaving, jiang2023structgpt}. It has shown successful performance on various tasks that require multiple data accesses to generate responses \cite{yang-etal-2018-hotpotqa, thorne-etal-2018-fever, ho2020constructing, aly-etal-2021-fact}, but is not powerful enough to solve the decision making task well in our experiments.

\section{Conclusions}
\label{sec:conclusions}

In this paper, we explored the capability of LLMs as a solution for decision making. 
We proposed the new decision-making task, \textbf{Decision QA}, which answers the best decision for a given complex decision-making question that requires considering both the business rules and business situation represented in a large database (in either RDB or GDB).
We built the benchmark for Decision QA, called \textbf{DQA}, by extracting 301 sets of a database (in both RDB and GDB), a question, and an answers(ground truth) from two popular video games imitating real business situations that require decision making.
We also proposed the new RAG technique called \textbf{PlanRAG}, which performs planning before retrieving and re-planning if the initial plan is not good enough. 
Through extensive experiments, we demonstrated that PlanRAG significantly outperforms the SOTA iterative RAG for the Decision QA task.

\section{Limitations}

In this paper, we explored the capability of LLM as a solution for decision making. However, our study still has several limitations.

First, in this study, we focused on Decision QA using graph database or relational database. Decision making based on other databases, such as a hybrid form of database and vector database, could be explored in future research.

Next, we proposed techniques from a high-level RAG technique perspective that should be considered when solving Decision QA. Therefore, we do not focus on the low-level methods for solving Decision QA in this paper. For example, creating a fine-tuned model that efficiently generates Cypher queries could be beneficial for solving Decision QA using GDB. These areas should also be addressed in future works.

Lastly, we designed our PlanRAG methodology and implemented our PlanRAG-LM using single LM, while some researches have suggest language model frameworks using multiple LMs. The effectiveness of the PlanRAG in multiple LMs framework is not a focus of this paper and is left as a further study.

\section{Ethical Considerations}
\label{sec:ethical-considerations}
Language models have a hallucination issue and can potentially generate biased answers. RAG methods we have discussed are known to mitigate these issues to some extent, but it does not imply that these issues do not occur.
Therefore, when applying our research to real-world applications, it is essential to closely examine whether the generated decisions are inferred based on hallucinated or biased knowledge.

Before constructing our benchmark and simulator from Europa Universalis IV and Victoria 3 games, we have considered end user license agreement (EULA)\footnote{\href{https://legal.paradoxplaza.com/eula?locale=en}{https://legal.paradoxplaza.com/eula?locale=en}} of their game publisher, Paradox Interactive.
Our benchmark and simulator correspond to gameplay and scripts of user generated content (UGC) in section 5 of EULA, and thus our content should be open-sourced. Therefore, we open our benchmark and simulator under the MIT license. Also, utilizing all icons that came from these games in our paper is classified as streaming Paradox Games in section 6 of EULA. According to EULA, we can freely use icons if our paper is not behind a paywall. 

Video games that we have used to construct \ourbenchmark{} describe historical situations. Therefore, our datasets, based on these games, include knowledge that contradicts contemporary common sense and might be aggressive towards certain groups.
For example, the correct answer that a specific nation should influence a particular region in the Locating scenario of our benchmark might be aggressive to specific nations or regions.
To avoid these issues, we anonymized the names of nations into three-letter codes rather than mentioning their names directly.
For example, instead of using the term \textit{Bahmanis Sultanate}\footnote{\href{https://en.wikipedia.org/wiki/Bahmani\_Sultanate}{https://en.wikipedia.org/wiki/Bahmani\_Sultanate}}, we employed the term \textit{BAH}, and instead of \textit{The Papel States}\footnote{\href{https://en.wikipedia.org/wiki/Papal\_States}{https://en.wikipedia.org/wiki/Papal\_States}}, we used \textit{PAP} as terminology.

\section*{Acknowledgements}
This work was supported by the National Research Foundation of Korea (NRF) grant funded by the Korea government(MSIT) (No. 2018R1A5A1060031, RS-2023-00281635) and Institute of Information \& communications Technology Planning \& Evaluation (IITP) grant funded by the Korea government(MSIT) (No. 2019-0-01267, GPU-based Ultrafast Multi-type Graph Database Engine SW).
\bibliography{anthology,custom}

\newpage

\appendix
\section{Appendix}
\subsection{Data collection} 
\label{appendix:data-coll}

To collect $D$, we first select the provided savefile from each game and extract related data by game savefile parser. The extracted data was stored in $D$ according to the DB schema of each scenario. Next, we generate $Q$ by applying components of $D$ into pre-defined question format for each scenario. 
To control the quality of our benchmark, we consider the following points:
\begin{itemize}
    \item For the Locating scenario, we create one problem for each country. As \textit{profit} is determined by $TP_{\textit{country}}$, we omit countries with low $TP_{\textit{country}}$ due to their minimal impact on decisions.
    \item For the Building scenario, we filtered problems which have multiple answers.
\end{itemize}

\subsection{Data instances}
\label{appendix:data-examples}
For both scenarios, the instances share the same business rules $R$. Some instances share the same database $D$, but not all use the same one. Here, each database $D$ has a different number of rows and different column values (e.g., LV, size of the building, current price of goods). 

In the Locating scenario, every instance has a different goal with a different pair of \{COUNTRY\_NODE\} and \{HOME\_NODE\} in the template of questions of \textit{“Where should I locate my merchant? My goal is maximizing} \{COUNTRY\_NODE\}\textit{’s profit on home node, }\{HOME\_NODE\}\textit{”}.

In the Building scenario, every instance has a different \{GOODS\} in the template of questions of \textit{“Which building ID should we increase a level by 5 to maximally decrease the market price of }\{GOODS\}\textit{?”}.

The instances in both scenarios are designed that LMs should retrieve different intermediate nodes and edges to calculate the values, such as \textit{TPR} for Locating and \textit{TS} for Building, required to make a best decision.

\subsection{Simulators for annotation}

Although applying every decision to real games and comparing the results is the most credible approach to annotate the best decision, it is impossible due to the following characteristics of games: (1) randomness and (2) not being open-sourced. 
Thus, we develop simulators for each scenario on \ourbenchmark{} to calculate decision results deterministically. 

Our simulator takes $D$ with \textit{decision} as input, calculate effects of decision, and returns updated $D$ by simulating based on business rules. To annotate answers for a question, we simulates all decision candidates.
For example, in Figure \ref{fig:example}, we locates a merchant on Doab and Ganges, our simulator calculates profit increments, 0.23 and 0.03 for each decisions, respectively. From these calculated results, we can annotate that Doab is the best decision for the question. Detailed algorithms are in Appendix \ref{appendix:sim-alg}

\subsection{Algorithm of simulators}
\label{appendix:sim-alg}
We provide an algorithm of simulator for each scenario for future research.
Before explaining each simulator, we introduce some additional symbols in Table \ref{tab:symbols}, while other symbols are described in Section \ref{sec:ourdataset}.

\begin{table}[htb!]
\centering
\footnotesize
\renewcommand{\arraystretch}{1.25}
\begin{tabular}{l|l}
\hline
\textbf{Symbol} & \textbf{Description} \\ \hline
\textbf{Locating} & \\
\textit{obj\_n} & The node we should calculate its flow \\
\textit{Nodes} & The set of nodes \\
\textit{Countries} & The set of countries \\
\textit{NodeCountry} & The set of NodeCountry relationships \\
\textit{Simulated\_n} & The set of nodes which are simulated \\ \hline
\textbf{Building} & \\
\textit{Goods} & The set of goods \\
\textit{Buildings} & The set of buildings\\
\textit{cycle\_cnt} & \# of hops that a building affects \\
$MIN(x,y)$ & Minimum value between $x$ and $y$\\
$AVG_Y(X)$ & For $y \in Y$, An average of the set $X(y)$\\ \hline
\end{tabular}
\caption{Summary of additional symbols.}
\label{tab:symbols}
\end{table}

\subsubsection{Locating simulator algorithm}
Algorithm \ref{alg:locating} describes the mechanism of Locating simulator. There are three different  functions in the Locating simulator: `trading\_power\_estimate', `flow\_estimation', and `find\_top\_n'.
\begin{itemize}
    \item \textbf{trading\_power\_estimate} takes the database $D$ as an input, calculates and updates $TP_\textit{country}(n,c)$ and $TP_\textit{total}(n)$ for all $n, c \in \textit{NodeCountry}$, and returns updated $D$.
    \item \textbf{flow\_estimation} takes the database $D$ and an objective node \textit{obj\_n} as inputs, calculates and updates \textit{flow(obj\_n, dest)} and IV(\textit{dest}) for all \textit{dest} such that $\textit{(obj\_n, dest)} \in F$ by using Eq. (\ref{eq:IV}), and returns updated $D$.
    \item \textbf{find\_top\_n} takes the database $D$ and \textit{Simulated\_n} as inputs and returns a node $n \in \textit{Nodes}-\textit{Simulated\_n}$ such that $ \forall_{\textit{n'} \in \textit{Nodes-Simulated\_n}} \textit{(n, n')} \notin F$. If $\textit{Nodes}-\textit{Simulated\_n} = \emptyset$, it returns \textit{NULL}.
\end{itemize}

\begin{algorithm}
\caption{Locating simulator}\label{alg:locating}
\footnotesize
\SetKwRepeat{Do}{do}{while}
\KwInput
{
    $D$ \quad\quad/* Database for Locating */
}
\KwOutput
{
    $D$ \quad\quad/* Database after calculating a \textit{profit} for every country in $D$. */ 
}
$D$$\gets $ trading\_power\_estimate($D$) /* Initialize $TP_{\textit{country}}(n,c)$ and $TP_{\textit{total}}(n)$ */\\
\textit{Simulated\_n}$\gets$\{\} \quad /* empty set. */\\
\Do{$\textit{Nodes}-\textit{Simulated\_n} \not=\emptyset$}
{
    \textit{obj\_n}$\gets$ find\_top\_n\textit{(D, Simulated\_n)} \\ 
    $D$ $\gets$ flow\_estimation\textit{(D, obj\_n)} \\ 
    \textit{Simulated\_n}.add\textit{(obj\_n)} \\ 
}
\For{$c, h \in \textit{Countries}$} {
    $\textit{profit}(c) \gets (LV(h,c)+IV(h,c)) \cdot \frac{TP_{\textit{country}}(h,c)}{TP_{\textit{total}}(h)}$ \quad\quad /* Eq. (\ref{eq:TPR}, \ref{eq:profit}) */ 
}
\textbf{return} $D$
\end{algorithm}

\subsubsection{Building simulator algorithm}

We provide the mechanism of Locating simulator in Algorithm \ref{alg:building}. We initialize $TS(g)$ as the sum of $MS(g,b)$ to avoid the situation where the price of the goods falls to the local minimum. We also limit maximum \textit{cycle\_cnt} as 10 to make our simulator more efficient.

\begin{algorithm}
\caption{Building simulator}\label{alg:building}
\footnotesize
\KwInput
{
    $D$ \quad\quad/* Database for Building */
}
\KwOutput
{
    $D$ \quad/* Database after calculating prices*/ 
}
\For{$g \in \textit{Goods}$}{
    $TD(g)\gets PD(g)$ \quad /* Eq. (\ref{eq:TD})*/ \\
    $TS(g)\gets 0$ \quad\quad /* Eq. (\ref{eq:TS})*/ 
}
\For{$g,b \in \textit{Dem}$} {
    $TD(g) \gets TD(g)+MD(g,b)$ \quad /* Eq. (\ref{eq:TD})*/ \\
}
\For{$g,b \in \textit{Sup}$}{
    $TS(g) \gets TS(g)+MS(g,b)$ /*TS initialize*/
} 
\textit{cycle\_cnt} $\gets 0$ \\
\While{$ \textit{cycle\_cnt} < 10$}
{
        \For{$g,b \in \textit{Dem}$}
        {
            $CI(g,b) \gets MD(g,b)*\frac{TS(g)}{TD(g)}$ \quad\quad\quad\quad /* From business rules in Figure \ref{fig:br-locatingNbuilding} (b). */
        }
        \For{$g \in \textit{Goods}$}
        {
            $TS(g)\gets 0$ \quad /* Eq. (\ref{eq:TS})*/
        }
        \For{$g, b \in \textit{Sup}$}
        {
            $CO(g, b) \gets MS(g,b) \cdot AVG_{{\{b'|(g, b') \in \textit{Dem}\}}}(MIN(1, \frac{CI(g,b')}{MD(g,b')}))$ \quad /* From business rules in Figure \ref{fig:br-locatingNbuilding} (b). */
        }
        \For{$g, b \in \textit{Sup}$} 
        {
            $TS(g) \gets TS(g)+CO(g,b)$ /* Eq. (\ref{eq:TS})*/ 
        }
        $\textit{cycle\_cnt} \gets \textit{cycle\_cnt} + 1$
}
\For{$g \in \textit{Goods}$}
{
    $CP(g)\gets BP(g)\cdot(1+0.75\cdot\frac{\textit{TD}(g)-\textit{TS}(g)}{\textit{max}(\textit{TD}(g), \textit{TS}(g))})$ 
    \quad\quad /* Eq. (\ref{eq:CP})*/ \\
}
\textbf{return} $D$
\end{algorithm}

\subsection{Business rules in \ourbenchmark{}}
\label{appendix:business-rules}
Figure \ref{fig:br-locatingNbuilding} shows the business rules $R$ used as LM's input in Locating and Building. 
These are textual descriptions of the contents in Section \ref{sec:ourdataset}.

\begin{figure}[htb!]
    \centering
    \footnotesize
    \fbox{\begin{minipage}{24em}
\textbf{(a)}
A "Trading node" has a "local\_value", "total\_power", "outgoing", "ingoing" and **whether it's inland**. A "Country" has a "name", "development" and a "home\_node" (home node). Between "Trading nodes“, there can be a directional edge [source]. It connects from a higher node to a lower node.  A "Country" can have a non-directional connection to a trading node. Each connection has a unique "base\_trading\_power" for each node. 
If a specific node is the home node of a country, that country earns profit from that node. The profit is proportional to the "local\_value" plus "ingoing" and the ratio of the country’s trading power and the total trading power of that node. i.e. (local\_value+ingoing)*(country\_trading\_power / total\_trading\_power)

If a specific node is source of a country's home node, the country moves a value to dest node, proportional to the ratio of the country's trading power to the node's total trading power and (local\_value + ingoing). In the dest node, the moved value is increased by 1.05 times and added to the ingoing. A "Merchant" belongs to a country and can be assigned to a specific trading node. A "Merchant" belonging to a trading node adds 2 to the trading power of the country's trading node and amplifies it by 1.05 times. *If one of the edges has an inland node, the added value changes to 2+max(development/3, 50). (optional)* If a specific trading node has more than one dest node, and a country that doesn't have that node as its home node places a "Merchant" on the Trading node, it can decide which dest node to move the "current\_value" to. That is, the country can move a "current\_value" proportional to its trading power to a specific dest node. If no merchant is placed, when there's more than one dest node, they lose the right to decide the direction. In other words, the country's "current\_value" proportional to its trading power flows out in proportion to other outflows to the dest nodes. If there's only one dest node, it doesn't matter.
\end{minipage}}
\fbox{\begin{minipage}{24em}
\textbf{(b)}
Goods have a "name", corresponding "code", "base\_price", "current\_price", and "pop\_demand". Buildings have a unique "id" and a "name" and "level" corresponding to their type. There exists a relation called Supply from Buildings to Goods. Supply has "max\_supply", "current\_output", and "level". The level here is the same as the level of the Building. Furthermore, max\_supply and level have a proportional relationship. There exists a relation called Demand from Goods to Buildings. Demand has "max\_demand", "current\_input", and "level". Also, max\_demand and level have a proportional relationship. The demand of Goods is defined as the Goods' "pop\_demand" plus sum of the "max\_demand" of all Demands connected to the Goods. The supply of Goods is defined by the sum of "current\_output" of all Supplies connected to Goods. The "current\_input" of Demand is determined by the ratio of connected Building's "max\_demand" to connected Goods' demand, and multiplied by the supply of Goods. The "current\_output" of Supply is determined by the average ratio of the connected Building's "current\_input" to connected Goods' "max\_demand", and multiplied by the "max\_supply" of Supply. The "current\_price" of Goods is determined by base\_price*(1+0.75(demand-supply)/max(demand,supply)).
\end{minipage}}
    \caption{Business rules for (a) Locating and (b) Building.}
    \label{fig:br-locatingNbuilding}
\end{figure}

\subsection{Prompts for DQA}
\label{appendix:prompt}

\subsubsection{Previous RAG-based LMs}
\label{appendix:prompt-rag}
\begin{figure}[htb!]
    \centering
    \includegraphics[width=\columnwidth]{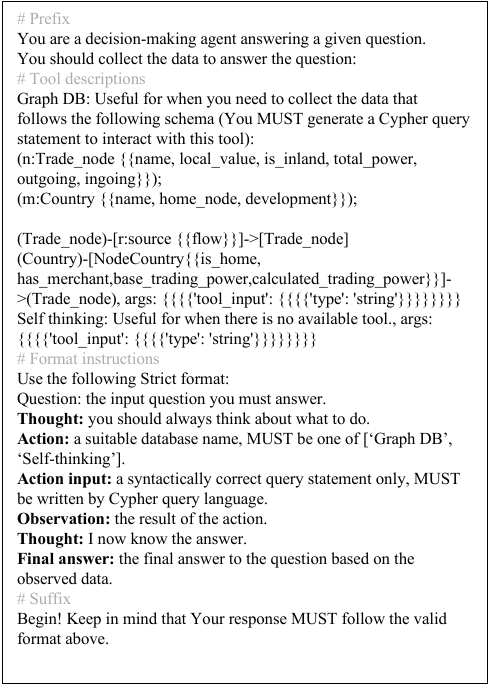}
    \caption{Retrieval prompt for a single-turn RAG technique on GDB case in Locating scenario.}
    \label{fig:SingleRAG-LM_retriever}
\end{figure}

\begin{figure}[htb!]
    \centering
    \includegraphics[width=\columnwidth]{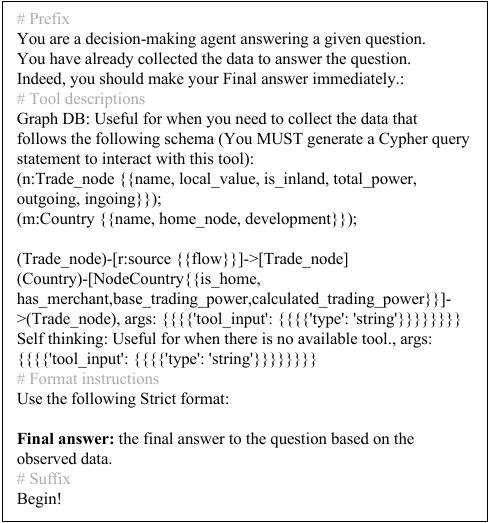}
    \caption{Answer generation prompt for a single-turn RAG technique on GDB case in Locating scenario.}
    \label{fig:SingleRAG-LM_answer}
\end{figure}

\begin{figure}[htb!]
    \centering
    \includegraphics[width=\columnwidth]{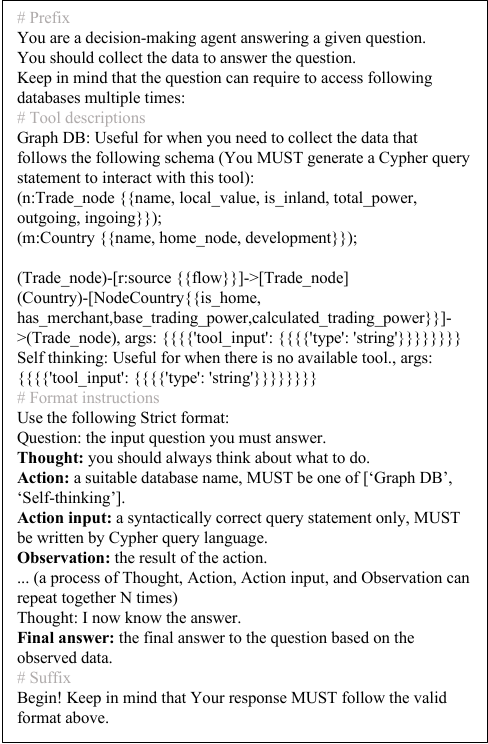}
    \caption{Prompt for an iterative RAG technique on GDB case in Locating scenario.}
    \label{fig:ReAct}
\end{figure}

Figure \ref{fig:SingleRAG-LM_retriever} and \ref{fig:SingleRAG-LM_answer} show an example of the prompts used for the single-turn RAG-based decision maker, SingleRAG-LM, in the Locating scenario with GDB setting.
Figure \ref{fig:ReAct} shows an example of the prompt used for the iterative RAG-based decision maker, IterRAG-LM, in the Locating scenario with GDB setting.
Those prompts are based on the structure of ReAct, `Thought'-`Action'-`Observation'. 

\subsubsection{PlanRAG-based LM}
\label{appendix:prompt-planrag}
\begin{figure}[htb!]
    \centering
    \includegraphics[width=\columnwidth]{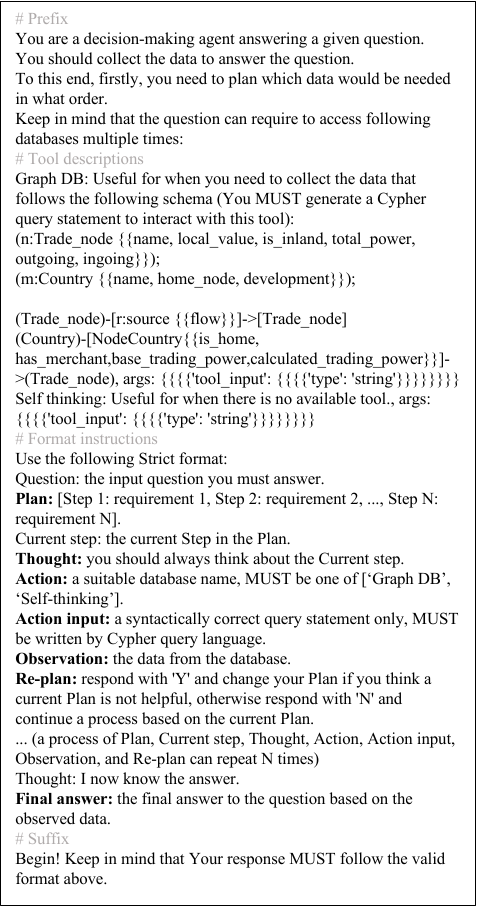}
    \caption{Prompt for PlanRAG technique on GDB case in Locating scenario.}
    \label{fig:PlanReAct}
\end{figure}

The prompt used to implement our PlanRAG-based decision maker, PlanRAG-LM, has a structure of `Plan’-`Thought’-`Action’-`Observation’-`Re-plan’.
Figure \ref{fig:PlanReAct} shows an example of the prompt for Locating scenario with GDB setting.
The prompt structure was designed empirically through experiments conducted on both DQA scenarios.

In the experiments, we also considered the following two variations of the prompt structure:
(1) Act without additional reasoning about a current step after planning (i.e., `Plan'-`Action'-`Observation'-`Re-plan'), 
(2) Reason about planning in advance (i.e., `Thought'-`Plan'-`Action'-`Observation'-`Re-plan').
The experiments were conducted by the same setup as in Section 5.1, using GPT-4 with zero temperature as base LMs in a zero-shot setting.

Table \ref{tab:results-prompts} shows the experimental results of the prompt structure of PlanRAG-LM, two aforementioned variations of that, and the ReAct prompt structure of IterRAG for 10\% of questions in each DQA scenario.
Here, in each scenario, we sampled the questions randomly at 10\% from each DB setting.
In the results, the prompt structure of PlanRAG-LM outperforms other baselines in both DQA scenarios.
Meanwhile, one PlanRAG-LM variation using the prompt structure of `Thought'-`Plan'-`Action'-`Observation'-`Re-plan' fails to demonstrate its effectiveness.
Another variation of PlanRAG-LM shows slightly better performance than the ReAct structure. 
The difference between these two results comes from whether to reason about the next action to take at each iteration, `Thought', or to act based on the plan established through the planning process, `Plan'.
The importance and effectiveness of the process `Thought' have been well discussed in several studies \cite{yao2022react, wei2022chain}.
Therefore, these results can show the significance of the planning process in decision-making tasks such as Decision QA.

\begin{table}[htb!]
\centering
\footnotesize
\renewcommand{\arraystretch}{1.25}
\begin{tabular}{lcc}
\hline
\textbf{Prompt structure} & \textbf{Locating} & \textbf{Building}  \\ \hline
\textbf{IterRAG-LM} & & \\
Thou-Act-Obs & 37.5 & 30 \\ \hline
\textbf{PlanRAG-LM} & & \\
Plan-Thou-Act-Obs-Replan & \textbf{57.5} & \textbf{50} \\ \hline
\textbf{PlanRAG-LM variations} & & \\
Plan-Act-Obs-Replan & 40 & 40 \\
Thou-Plan-Act-Obs-Replan & 27.5 & 30 \\ \hline
\end{tabular}
\caption{Accuracy(\%) of LMs for DQA by prompt structure. `Thou', `Act', and `Obs' means `Thought', `Action', and `Observation', respectively.}
\label{tab:results-prompts}
\end{table}

\subsection{Experimenting with other models}
\label{appendix:open-model}
In this section, we implement IterRAG-LM and PlanRAG-LM by four different models: (1) GPT-3.5\footnote{gpt-3.5-turbo-0125, which is the latest gpt-3.5 model.} (2) Llama 2 (70B), (3) Llama 2 (13B) \cite{touvron2023llama}, and (4) Phi-2  \cite{javaheripi2023phi}.
All experiments are conducted on a single machine equipped with eight Nvidia A100 (80GB) GPUs. To accelerate inference speed, we utilize vLLM \cite{kwon2023efficient} library for open models inference. We set temperature to zero and 0.1 for GPT-3.5-turbo and other open models, respectively. Other settings are consistent with those described in Section \ref{sec:experiments}.
We provide results of PlanRAG-LM and IterRAG-LM by four models in Table \ref{tab:open-model-results}. 
In the result, PlanRAG-LM and IterRAG-LM by Llama-2 and Phi-2 models cannot solve any problems in DQA. By GPT-3.5, IterRAG-LM shows batter performance rather than PlanRAG-LM. It is because the prompt of PlanRAG is too complex for GPT-3.5 to understand instructions and generate proper answers.

\begin{table}[htb!]
\renewcommand{\arraystretch}{1.25}
    \footnotesize
    \centering
    \begin{tabular}{lcccc}
    \hline
    & \multicolumn{2}{c}{\textbf{Locating}} & \multicolumn{2}{c}{\textbf{Building}} \\ \cline{2-5} 
    \textbf{Models} & RDB & GDB & RDB & GDB \\ 
    \hline
    \textbf{GPT-3.5} & & & &\\
    IterRAG-LM & \textbf{8.0} & 2.5 & \textbf{22.8} & \textbf{3.96} \\
    PlanRAG-LM & 0 & \textbf{4.0} & 1.0 & 1.0 \\
    \textbf{Llama 2 (70B)} & & & &\\
    IterRAG & 0 & 0 & 0 & 0 \\
    PlanRAG & 0 & 0 & 0 & 0 \\
    \textbf{Llama 2 (13B)} & & & &\\
    IterRAG & 0 & 0 & 0 & 0 \\
    PlanRAG & 0 & 0 & 0 & 0 \\
    \textbf{Phi-2} & & & &\\
    IterRAG & 0 & 0 & 0 & 0 \\
    PlanRAG & 0 & 0 & 0 & 0 \\ 
    \hline
    \end{tabular}
    \caption{Accuracy(\%) of IterRAG-LM and PlanRAG-LM using several models.}
    \label{tab:open-model-results}
\end{table}

\subsection{Re-planning cases}
\label{appendix:replan-case}

Table \ref{tab:replan-example} presents statistics and examples of re-plannings conducted by PlanRAG-LM.
We categorized all re-planning cases into three groups: \textbf{(1) Increase}, \textbf{(2) Same}, and \textbf{(3) Decrease}, where “Increase” means the number of steps increases after re-planning compared to the original plan, “Same” means the number of steps is the same with the number of steps of the original plan, and “Decrease” means the number of steps decreases after re-planning.

Each category is further divided into the following sub-categories:
\begin{itemize}[nosep]
    \item \textbf{Re-order} includes cases where a sequence of steps is arranged.
    \item \textbf{Replace} includes cases where some steps are substituted with new steps.
    \item \textbf{Change target} includes cases where targets of actions, such as lookup or calculation, are changed.
    \item \textbf{Add look-up} includes cases where new lookup actions are added to an original plan.
    \item \textbf{Add calculation} includes cases where new calculation actions are added to an original plan.
    \item \textbf{Add both actions} includes cases where both the lookup and the calculation actions are added to an original plan by a single re-planning process.
    \item \textbf{Divide to sub-steps} includes cases where a single step of an original plan is divided into sub-steps by breaking it down into detailed actions.
    \item \textbf{Delete} includes cases where some steps are removed from an original plan.
    \item \textbf{Merge} includes cases where some steps are summarized or merged into a single step.
\end{itemize}
As shown in Table \ref{tab:replan-analysis}, since re-planning is done more than twice for some questions, they can belong to several categories.
In the Locating scenario, PlanRAG-LM primarily performs re-planning of "Add look-up".
In the case of Building, it is observed that "Divide to sub-steps" is the predominant strategy for re-planning across most cases.

\begin{table*}[htb!]
\centering
\includegraphics[width=\linewidth]{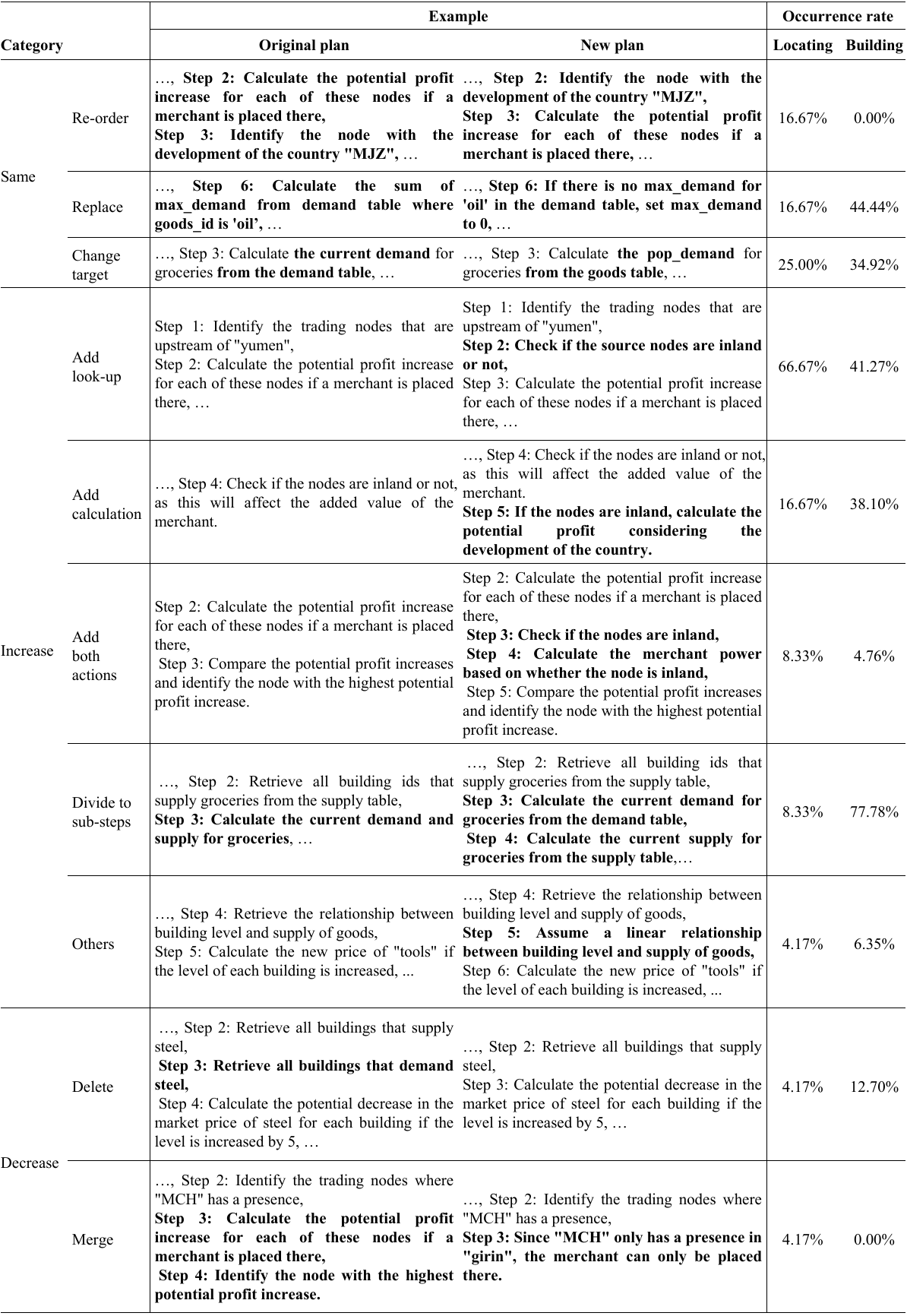}
\caption{Category-wise statistics and examples of re-planning cases. The occurrence rate indicates the proportion of re-planning cases within a specific category relative to the total number of questions that re-plannings are conducted.}
\label{tab:replan-example}
\end{table*}

\end{document}